\documentclass[pdflatex,sn-mathphys-num]{sn-jnl}

\usepackage{tabularx}
\usepackage{float} 
\usepackage{graphicx}%
\usepackage{multirow}%
\usepackage{makecell}
\usepackage{amsmath,amssymb,amsfonts}%
\usepackage{amsthm}%
\usepackage{array}
\usepackage{mathrsfs}%
\usepackage[title]{appendix}%
\usepackage{textcomp}%
\usepackage{manyfoot}%
\usepackage{booktabs}%
\usepackage{algorithm}%
\usepackage{algorithmicx}%
\usepackage{algpseudocode}%
\usepackage{listings}%
\usepackage{geometry}  
\usepackage{lineno}
\usepackage{minitoc}
\usepackage{titletoc}
\usepackage{fancyhdr}
\usepackage[table,xcdraw]{xcolor}
\usepackage{tikz,lipsum,lmodern}
\usepackage[most]{tcolorbox}
\usepackage{pdfpages}
\usepackage{hyperref} 
\usepackage{gensymb}
\tcbuselibrary{listings}
\usepackage{listingsutf8}
\usepackage{lineno}
\usepackage{bibunits}
\usepackage{chngcntr}
\usepackage{hyperref}
\usepackage{cleveref}

\usepackage[normalem]{ulem}
\newcommand{\stkout}[1]{\ifmmode\text{\sout{\ensuremath{#1}}}\else\sout{#1}\fi}

\theoremstyle{thmstyleone}%
\theoremstyle{thmstyletwo}%

\theoremstyle{thmstylethree}%

\begin{document}

\setlength{\bibsep}{0.5em}

\Crefname{figure}{Fig.}{Figs.}

\renewcommand\linenumberfont{\normalfont\scriptsize}
\setlength\linenumbersep{25pt}

\title[Owl·AuraID 1.0]{Owl·AuraID 1.0: An Intelligent System for Autonomous Scientific Instrumentation and Scientific Data Analysis}


\author[1,2]{\fnm{Han} \sur{Deng}}\email{dh025@ie.cuhk.edu.hk}
\equalcont{These authors contributed equally to this work.}

\author[1,3]{\fnm{Anqi} \sur{Zou}}\email{bigben@mail.dlut.edu.cn}
\equalcont{These authors contributed equally to this work.}

\author[1,2]{\fnm{Hanling} \sur{Zhang}}\email{zh025
@ie.cuhk.edu.hk}
\equalcont{These authors contributed equally to this work.}

\author*[2]{\fnm{Ben} \sur{Fei}}\email{benfei@cuhk.edu.hk}

\author[1]{\fnm{Chengyu} \sur{Zhang}}

\author[1,3]{\fnm{Haobo} \sur{Wang}}

\author[1,3]{\fnm{Xinru} \sur{Guo}}

\author[1,3]{\fnm{Zhenyu} \sur{Li}}

\author[1,3]{\fnm{Xuzhu} \sur{Wang}}

\author[4]{\fnm{Peng} \sur{Yang}}

\author[4]{\fnm{Fujian} \sur{Zhang}}

\author[5]{\fnm{Weiyu} \sur{Guo}}\email{guoweiyu96@gmail.com}

\author[4]{\fnm{Xiaohong} \sur{Shao}}\email{shaoxh@szlab.ac.cn}

\author[5]{\fnm{Zhaoyang} \sur{Liu}}\email{zyliumy@gmail.com}

\author[2]{\fnm{Shixiang} \sur{Tang}}\email{shixiangtang@cuhk.edu.hk}

\author*[1,3]{\fnm{Zhihui} \sur{Wang}}\email{zhwang@dlut.edu.cn}

\author*[1,2]{\fnm{Wanli} \sur{Ouyang}}\email{wlouyang@ie.cuhk.edu.hk}

\affil*[1]{\orgname{Shenzhen Loop Area Institute}, \orgaddress{\city{Shenzhen}, \country{China}}}

\affil[2]{\orgname{The Chinese University of Hong Kong}, \orgaddress{\city{Hong Kong}}}

\affil[3]{\orgname{Dalian University of Technology}, \orgaddress{\city{Dalian}, \country{China}}}

\affil[4]{\orgname{Suzhou Laboratory}, \orgaddress{\city{Suzhou}, \country{China}}}

\affil[5]{\orgname{The Hong Kong University of Science and Technology}, \orgaddress{\city{Hong Kong}}}

\abstract{
    Scientific discovery increasingly depends on high-throughput experiments and complex characterization instruments, yet the practical automation of scientific characterization remains limited by proprietary graphical user interfaces (GUIs) that lack programmable APIs. Existing API- or command-line-based automation methods often fail to capture the visual feedback, software state, and tacit heuristics required in real characterization workflows, while many self-driving laboratory systems rely on pre-integrated devices or fixed software stacks that are difficult to generalize across heterogeneous equipment. Here, we present \textbf{Owl-AuraID} (Aura AI for Identification), a software--hardware collaborative embodied agent system for autonomous scientific instrumentation and scientific data analysis. Owl-AuraID adopts a GUI-native computer-use paradigm that allows agents to operate instrument control software through the same interfaces used by human experts, and organizes reusable capability through a skill-centric framework. In this framework, \textbf{Type-1 skills} encode expert-demonstrated GUI operational procedures for instrument control, and \textbf{Type-2 skills} encapsulate analytical scripts for post-acquisition data interpretation. Together, these skills connect physical sample handling, instrument software operation, and scientific analysis into an end-to-end characterization workflow. We report broad laboratory coverage across ten categories of precision scientific instruments and representative workflows spanning multimodal spectral analysis, multiscale microscopic imaging, elemental mapping, tomographic reconstruction, and crystallographic structure analysis, together with additional analytical support for modalities such as Fourier-transform infrared spectroscopy (FTIR), Nuclear Magnetic Resonance Spectroscopy (NMR), Atomic Force Microscopy (AFM), and Thermogravimetric Analysis (TGA). Overall, Owl-AuraID provides a practical and extensible foundation for autonomous scientific characterization in heterogeneous real-world laboratories and illustrates a path toward continually evolving laboratory intelligence through reusable operational and analytical skills.
    \textbf{Code:} \url{https://github.com/OpenOwlab/AuraID}.
}

\maketitle

\section{Introduction}

Recent advances in AI for Science (AI4Science) have dramatically expanded our ability to generate scientific hypotheses and explore vast candidate spaces in biology, chemistry, and materials science. Breakthroughs such as AlphaFold~\cite{AlphaFold2021}, AlphaFold~3~\cite{AlphaFold3_2024}, and large-scale materials discovery exemplified by GNoME~\cite{GNoME2023} have demonstrated that machine intelligence can capture complex physical and biological constraints at unprecedented scale. Yet the experimental side of scientific discovery has not progressed at the same pace. In practice, the validation of computational hypotheses still depends on labor-intensive laboratory workflows, creating a widening gap between rapid \textit{in silico} discovery and slow wet-lab verification.

A major bottleneck in this gap lies in scientific characterization. High-precision characterization is essential for connecting experimental outcomes to material structures, chemical states, and functional properties, but it remains one of the least standardized stages of laboratory automation~\cite{SDLReview2025}. Compared with robotic synthesis and reaction optimization~\cite{AIDrivenAutonomousLab2025,MobileRoboticProcessChemist}, characterization is substantially harder to automate in real laboratories because it depends on heterogeneous instruments, vendor-specific software, and expert-driven interpretation. In many practical settings, critical operational knowledge is not exposed through programmable interfaces, but instead embedded in graphical control software, undocumented interaction procedures, and the tacit heuristics of experienced operators. As a result, high-precision characterization remains one of the most difficult barriers to practical laboratory autonomy.

This challenge is not merely a hardware integration problem, but also a software interaction problem. Existing autonomous laboratory systems have achieved impressive progress in robotic synthesis, reaction optimization, and closed-loop experimentation~\cite{SDLReview2025}, including integrated materials platforms~\cite{ALab2023,AutonomousLabNovelMaterials,CederAIInAction2026}, mobile laboratory agents~\cite{MobileRoboticChemist2020}, and workflow orchestration stacks~\cite{AlabOS2024,UniLabOS2025}. However, most current approaches still rely on pre-integrated device APIs, custom middleware, or tightly engineered hardware-software stacks, a pattern also reflected in programmatic scientific agents~\cite{Coscientist2023,ChemCrow2024}. Such designs are effective in purpose-built automation settings, but are considerably harder to deploy in real-world laboratories populated by heterogeneous, vendor-specific, and often closed-source instruments that lack programmable interfaces. For characterization in particular, many key decisions are made through visual interaction with spectra, microscopy images, parameter panels, and real-time instrument feedback. These properties make GUI-native interaction a more natural foundation than API-only abstraction for practical characterization automation.

In this work, we argue that scientific instruments should be regarded as specialized embodied AI agents, for which scientific software serves as the first-class operating environment. 
Rather than assuming that laboratory intelligence must be built exclusively on top of pre-defined APIs---or that proprietary interfaces can always be wrapped into command-line tool schemas~\cite{CLIAnything2025,OpenCLI2025}---we advocate a GUI-native paradigm in which agents perceive and manipulate the same interfaces used by human experts, consistent with progress in computer-use agents for real desktop and mobile software~\cite{OSWorld2024,AndroidWorld2024,team2026kimi,claude4_6,gpt54}. 
To realize this paradigm in practice, we introduce \textbf{Owl-AuraID} (Aura AI for Identification), a software-hardware collaborative embodied system built around a scientific agent runtime and a reusable skill-centric capability framework. Owl-AuraID adapts the agent-runtime paradigm of modern coding agents to scientific characterization, providing a conversation-first interface for non-programmer scientific users while retaining workspace access, command execution, and autonomous tool use under the hood. Its core capability is organized around reusable scientific skills: \textit{Type-1 skills}, which encode expert GUI operational procedures for proprietary instrument software, and \textit{Type-2 skills}, which encapsulate analytical scripts for transforming raw characterization outputs into scientific results. By treating skills as the primary unit of accumulated capability, Owl-AuraID turns one-off successful interactions into reusable, extensible procedures that can be invoked, refined, and composed across tasks, thereby supporting continual capability accumulation over time.

Built on this foundation, Owl-AuraID connects physical sample handling with robots, GUI-based instrument control, and scientific data interpretation into an end-to-end characterization workflow. The system supports a broad range of precision characterization scenarios, spanning multimodal spectral analysis, multi-scale microscopic imaging, elemental mapping, crystallographic structure analysis, and related data-processing tasks. In this way, Owl-AuraID is not merely a GUI automation system for scientific software, but a skill-centric scientific agent platform that links embodied execution and digital analysis in a unified and extensible loop.

In summary, this work makes four main contributions:
\begin{itemize}
    \item We formulate autonomous scientific characterization in real laboratories as a GUI-based computer-use problem, identifying the lack of open APIs and the prevalence of heterogeneous vendor software as a central obstacle to practical laboratory autonomy.
    \item We present Owl-AuraID, a software-hardware collaborative scientific agent system that adapts the agent-runtime paradigm to characterization workflows and enables end-to-end operation across heterogeneous scientific instruments without requiring pre-defined tool interfaces.
    \item We propose a reusable skill-centric framework in which \textit{Type-1 GUI operational skills} and \textit{Type-2 analytical script skills} serve as the primary units of capability accumulation, allowing expert procedures and heuristics to be encoded, reused, and composed across tasks.
    \item We demonstrate a unified characterization-analysis loop that connects embodied manipulation, instrument software operation, and scientific data interpretation across diverse characterization scenarios, supporting continual capability expansion in autonomous laboratory systems.
\end{itemize}

\begin{figure}[htbp]
    \centering
    \includegraphics[width=1\textwidth]{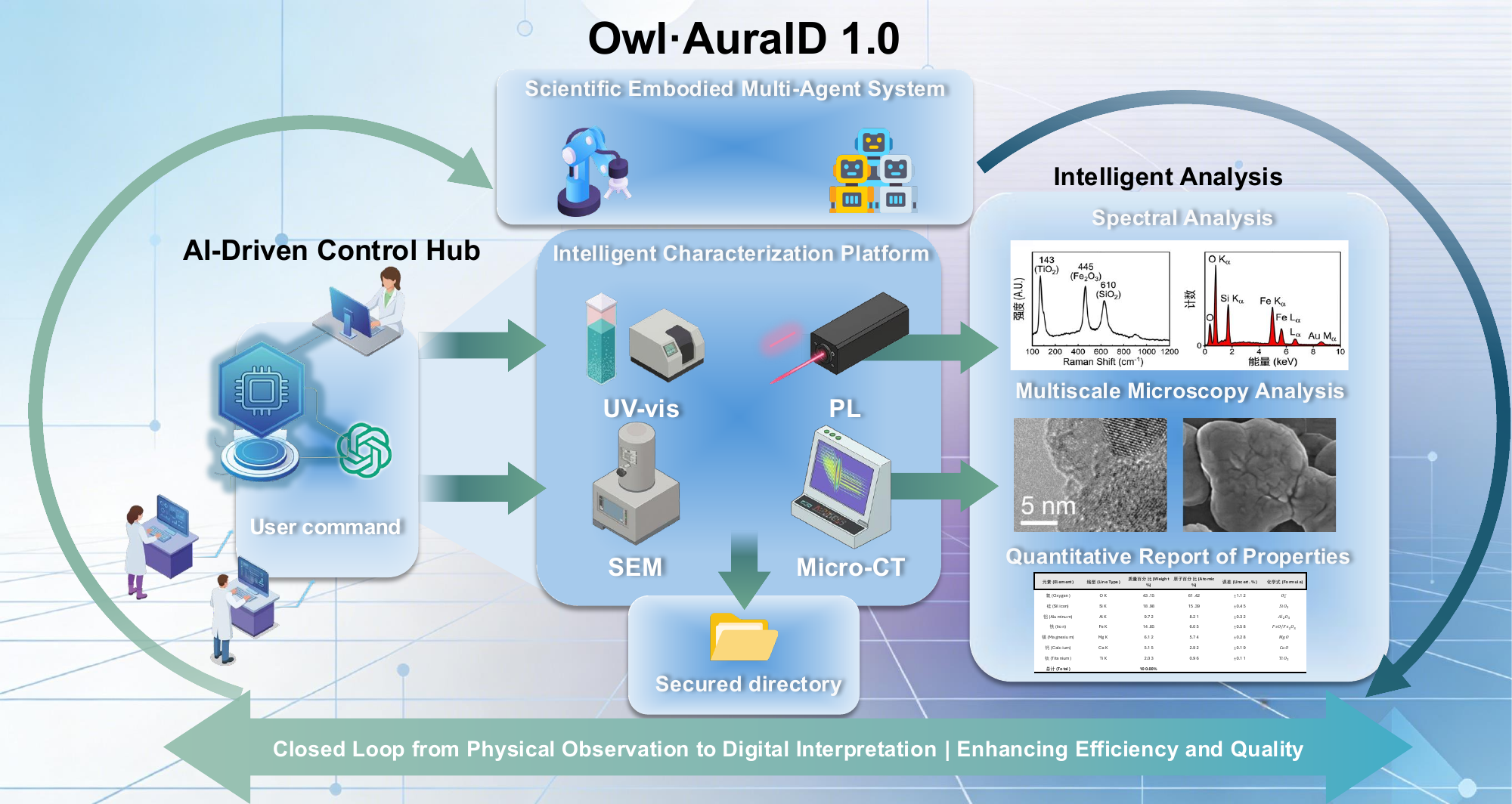} 
    \caption{\textbf{Overview of the Owl-AuraID 1.0 architecture for autonomous scientific characterization.}
    User command is translated and executed by an embodied multi-agent system managing a suite of precision characterization instruments (UV-vis, PL, SEM, Micro-CT). 
    The platform achieves scientific data acquisition and autonomous multi-modal analytics (multiscale imaging, spectral analysis, and properties quantification), establishing an efficient loop for rapid characterization-analysis cycles.
    }
    \label{fig:overview}
\end{figure}

\section{Related Work}

\begin{table*}[t]
    \centering
    \caption{Comparison of representative systems related to laboratory automation and characterization-oriented workflows.}
    \label{tab:comparison_lab_systems}
    \scriptsize
    \setlength{\tabcolsep}{3pt}
    \renewcommand{\arraystretch}{1.15}
    \begin{tabularx}{\textwidth}{
      >{\raggedright\arraybackslash}p{2.5cm}
      >{\raggedright\arraybackslash}X
      >{\centering\arraybackslash}p{0.9cm}
      >{\centering\arraybackslash}p{2.3cm}
      >{\centering\arraybackslash}p{1.0cm}
      >{\centering\arraybackslash}p{1.0cm}
      >{\centering\arraybackslash}p{1.1cm}
    }
    \toprule
    \multirow{2}{*}{Category} & \multirow{2}{*}{Method} & \multicolumn{5}{c}{Characterization-Oriented Workflow Capabilities} \\
    \cmidrule(lr){3-7}
    & & Load & Instrument SW Control & Analysis & Return & Evolve \\
    \midrule
    
    \multirow{2}{*}{\makecell[l]{AI-assisted\\experimentation}}
    & InternSpect~\cite{InternSpectScientificDiscoveryPlatform}
    & N/A & N/A & Yes & N/A & No \\
    & LabOS / LabClaw~\cite{LabOS2025,LabClaw2026}
    & N/A & N/A & Yes & N/A & Yes \\
    \midrule
    
    \multirow{7}{*}{\makecell[l]{Automated\\laboratory}}
    & Venus series~\cite{VenusSeries}
    & Yes & Robotic GUI actuation & Yes & Yes & No \\
    & Mobile Robotic Chemist~\cite{MobileRoboticChemist2020}
    & Yes & Robotic GUI actuation & Yes & Yes & No \\
    & LG Chemistry Autonomous Lab~\cite{LGChemistryAutonomousLab}
    & Yes & Robotic GUI actuation & N/A & Yes & No \\
    & A-Lab~\cite{ALab2023}
    & Yes & Robotic GUI actuation & N/A & Yes & No \\
    & Chemspeed~\cite{ChemspeedPlatform}
    & Yes & Robotic GUI actuation & N/A & Yes & No \\
    & LUMI-lab~\cite{LUMILab}
    & Yes & Robotic GUI actuation & Yes & Yes & No \\
    & Uni-Lab-OS~\cite{UniLabOS2025}
    & Yes & Tool/API & Yes & Yes & No \\
    \midrule
    
    \makecell[l]{Fully intelligent\\laboratory}
    & Owl-AuraID (ours)
    & Yes & GUI-native agent\textsuperscript{a} & Yes & Yes & Yes\textsuperscript{b} \\
    \bottomrule
    \end{tabularx}
    
    \vspace{2pt}
    \raggedright\scriptsize \textsuperscript{a} Supports arbitrary scientific characterization software without predefined tool interfaces. \\
    \raggedright\scriptsize \textsuperscript{b} ``Evolve'' denotes whether the system can accumulate reusable operational or analytical capabilities across tasks.
\end{table*}

\subsection{AI for Scientific Discovery and Self-Driving Laboratories}

AI4Science as fundamentally reshaped discovery across biology, chemistry, and materials science.
Landmark systems, notably AlphaFold~\cite{AlphaFold2021, AlphaFold3_2024}, have demonstrated the capacity of deep learning to resolve complex biomolecular structures and interactions with unprecedented fidelity. Similarly, in materials science, large-scale deep learning and active-learning frameworks like GNoME~\cite{GNoME2023} have dramatically expanded the known space of stable materials. However, these computational leaps expose a widening gap: while the generation of scientific hypotheses has accelerated exponentially, the physical tasks of synthesis, characterization, and validation remain a bottleneck.

To bridge this gap, the scientific community is increasingly pivoting toward \emph{self-driving laboratories} (SDLs)—closed-loop systems that integrate automated experimentation with computational decision-making. Recent syntheses position SDLs as the infrastructural cornerstone of next-generation discovery, unifying robotics, workflow orchestration, and machine learning into cohesive, high-throughput architectures \cite{SDLReview2025}. In chemistry, these autonomous platforms have already accelerated reaction screening and iterative hypothesis testing by coupling robotic execution with model-based planning \cite{AIDrivenAutonomousLab2025}. Parallel advances in materials science, exemplified by A-Lab \cite{ALab2023} and integrated synthesis systems \cite{AutonomousLabNovelMaterials, CederAIInAction2026}, demonstrate that the tight convergence of robotic preparation, characterization, and active learning can drastically speed up the materials discovery cycle.

Going beyond fixed workcells, the mobile robotic chemist~\cite{MobileRoboticChemist2020} demonstrated that autonomous agents can navigate conventional laboratories to execute complex experimental campaigns. Subsequent advancements in mobile process chemistry~\cite{MobileRoboticProcessChemist} have further generalized this paradigm across broader, process-oriented tasks. At the architectural layer, frameworks such as AlabOS~\cite{AlabOS2024} and UniLabOS~\cite{UniLabOS2025} underscore the necessity of AI-native operating systems to orchestrate heterogeneous devices, analytical modules, and optimization loops. More recently, XR-assisted and multimodal ``co-scientist'' systems—notably LabOS~\cite{LabOS2025} and STELLA~\cite{STELLA2025}—have pushed this frontier toward highly interactive and adaptive laboratory intelligence.

Despite these advances, most current SDL systems rely on a degree of pre-engineered integration—such as standardized device APIs, bespoke middleware, or automation-centric laboratory redesigns. While feasible in purpose-built environments, these assumptions falter in real-world characterization laboratories. Here, instruments remain highly heterogeneous, vendor software is predominantly closed-source, and operational control frequently depends on manual visual interaction with legacy interfaces. Unlike synthesis or liquid-handling workflows, characterization is characterized by a lack of software standardization and a heavy reliance on proprietary, human-centric control programs. Consequently, characterization persists as one of the most fragmented and challenging bottlenecks in achieving true laboratory autonomy.

\subsection{LLM Agents for Scientific Experimentation}

Another emerging direction explores the use of large language models (LLMs) as scientific agents for experiment planning, tool invocation, and knowledge integration. 
Systems such as Coscientist~\cite{Coscientist2023} demonstrated that LLMs can assist with experimental design and execution when connected to programmable laboratory endpoints and documented automation interfaces.
ChemCrow~\cite{ChemCrow2024} further showed that tool-augmented language agents can solve chemistry tasks by orchestrating external models, software tools, and domain knowledge resources. 
AI Scientist~\cite{lu2026towards} introduces a transformative framework capable of managing the entire scientific discovery process—from initial hypothesis generation and code-based experimentation to automated manuscript preparation and peer review.
More broadly, recent studies on tool-using scientific agents have investigated when and how external tools improve chemistry reasoning and problem solving~\cite{ToolingOrNotChemistryAgents}.

These works illustrate the growing promise of LLMs as scientific reasoning engines and workflow coordinators. However, most existing approaches are built around environments in which relevant tools are already programmatically accessible through APIs, scripts, databases, or explicitly defined tool schemas. They are therefore better suited to scientific reasoning \emph{over} available tools than to autonomous operation \emph{of} closed, GUI-centric scientific software. This distinction is especially important for characterization settings, where key operations frequently must be carried out through proprietary graphical interfaces rather than open programmatic endpoints.

\subsection{GUI Agents for Real-World Software Operation}

In parallel, GUI agents, also known as Computer-Use Agents (CUAs), have made rapid progress as a general framework for operating software directly through visual interfaces. Benchmarks such as OSWorld~\cite{OSWorld2024} and
AndroidWorld~\cite{AndroidWorld2024} has provided realistic environments for evaluating agents in open-ended desktop and mobile tasks, helping drive progress in visual grounding, action planning, and long-horizon software interaction. Representative open models include UI-TARS~\cite{UITARS2025} and Kimi K2.5~\cite{team2026kimi}, while closed models such as Claude Sonnet 4.6~\cite{claude4_6}, GPT-5.4~\cite{gpt54}, and Seed-1.8~\cite{seed1_8} have demonstrated strong performance in GUI grounding and computer-use tasks. 
In particular, recent results on OSWorld-Verified suggest that state-of-the-art GUI agents are approaching or surpassing human-level performance in some software operation benchmarks.

These developments indicate that GUI-native interaction is becoming a viable foundation for real-world software automation. However, most current GUI-agent evaluations focus on general-purpose desktop or mobile applications, such as productivity tools, websites, and consumer operating environments. 
Scientific instrumentation differs in several important respects: interfaces are often vendor-specific and weakly standardized; workflows are strongly stateful and long-horizon; visual cues must be interpreted in relation to measurement quality and experiment status; and mistakes may waste samples, time, or instrument resources. 
More importantly, scientific instrument control is rarely an isolated software task, but is tightly coupled to physical experiment execution and downstream scientific interpretation. As a result, the direct application of general-purpose GUI agents to characterization laboratories remains relatively underexplored.

\subsection{Why GUI-Native Interaction Matters for Scientific Characterization}

An alternative line of work seeks to convert existing software into command-line or structured tool interfaces for LLMs. Systems such as CLI-Anything~\cite{CLIAnything2025} and OpenCLI~\cite{OpenCLI2025} have shown that web and desktop applications can sometimes be wrapped into LLM-friendly CLI schemas, enabling programmatic interaction without direct GUI operation.
This strategy is attractive when software is open, interface boundaries are stable, and the interaction logic can be cleanly abstracted into tool calls.
In such settings, API- or CLI-first interaction may indeed be preferable for reliability and efficiency.

For scientific characterization software, however, these assumptions often do not hold. 
First, many instrument control packages are proprietary and closed-source, with no publicly documented APIs, making reliable CLI conversion technically difficult and sometimes legally ambiguous. Second, building and maintaining such wrappers requires substantial software engineering effort, which is rarely available in ordinary laboratory settings.
Third, characterization is intrinsically visual: operators make decisions based on live images, spectral previews, parameter panels, reconstruction interfaces, and real-time feedback displayed in the GUI. A pure CLI abstraction cannot fully preserve this visual analytical context or the tacit decision-making process associated with it.

For these reasons, GUI-native interaction is not merely a fallback for missing APIs, but a natural operating paradigm for heterogeneous scientific instrumentation in real laboratories. 
Our work builds on recent progress in autonomous laboratories, LLM-based scientific agents, and GUI automation, while addressing a gap not fully covered by prior systems: enabling AI agents to directly operate real-world scientific characterization software and to connect physical execution, instrument software control, and scientific data analysis within a unified workflow.

Overall, prior work leaves three gaps insufficiently addressed for real-world scientific characterization: 
(1) most autonomous laboratory systems rely on pre-integrated APIs or fixed automation stacks; 
(2) most LLM scientific agents assume that relevant tools are already programmatically accessible; and
(3) most GUI agents are evaluated in general-purpose software environments rather than proprietary, safety-sensitive scientific instrumentation. 
Our work addresses these gaps through a GUI-native scientific agent framework that integrates embodied execution, instrument software operation, and scientific data analysis across heterogeneous characterization workflows.

\section{Method}

\subsection{Design Motivation}

Scientific characterization workflows present a distinctive systems challenge for AI agents. On the one hand, many tasks are inherently \emph{agentic}: they require multi-step planning, tool use, intermediate inspection, and iterative correction rather than single-turn question answering. On the other hand, the operational environment of characterization is highly heterogeneous. A practical system must bridge scriptable data analysis, proprietary GUI-based instrument software, and non-programmer human users within the same workflow.

This setting exposes a gap in existing agent paradigms. Pure conversational assistants lack the ability to act over files, commands, and software environments. In contrast, coding agents such as Claude Code- or CodeX-style systems provide strong command-line and workspace execution abilities, but their default interaction model is centered around code manipulation in developer-oriented environments. Such a purely CLI-centric paradigm is poorly aligned with scientific instrument scenarios, where critical steps often take place inside closed, GUI-driven software stacks and where end users are experimental researchers rather than professional programmers.

AuraID is designed to address this gap. Built on top of InnoClaw~\cite{InnoClaw2026}, it extends the agent-runtime paradigm beyond software engineering and adapts it to scientific characterization. Our central design principle is that a useful laboratory agent should not be framed as a monolithic model application, but as a \emph{skill-centric agent platform}: a system that can reason over tasks, act over both workspace and GUI environments, and accumulate reusable procedural knowledge from expert interaction.

Accordingly, the method is organized around three components:
\begin{enumerate}
    \item An agent runtime platform tailored to scientific users and mixed execution environments;
    \item A skill system that serves as the primary unit of accumulated capability;
\end{enumerate}

\subsection{AuraID as a Scientific Agent Runtime}

\subsubsection{From Coding Agent to Scientific Agent}

AuraID is developed as a secondary system on top of InnoClaw~\cite{InnoClaw2026}. At the architectural level, it inherits the essential properties of modern agentic coding systems: workspace access, command execution, model-driven planning, and an autonomous agent loop that iteratively acts until task completion. This agent-runtime structure is crucial because scientific workflows are rarely solvable in a single model response. Even simple requests such as analyzing a spectrum, reproducing a reconstruction pipeline, or exporting instrument results typically require multiple rounds of file inspection, tool invocation, parameter adjustment, and result verification.

However, directly applying a coding-agent paradigm to laboratory environments is insufficient. Systems such as Claude Code or CodeX are highly effective in developer workflows because they assume a command-line-native workspace and users who are comfortable inspecting, editing, and debugging code artifacts. Scientific characterization differs in two important respects.

First, many critical operations are not exposed through scripts or open APIs, but through proprietary software interfaces operated primarily via GUI. In such settings, a purely CLI-based agent cannot directly reach important parts of the workflow. Second, the target users of AuraID are often instrument operators, materials researchers, and characterization scientists, whose goals are experimental and analytical rather than code-centric. Requiring such users to manually modify source code would introduce unnecessary cognitive and operational burden.

For this reason, AuraID retains the underlying execution power of coding agents, while redesigning the interaction layer for scientific use.

\subsubsection{Conversation-First Interaction over a Tool-Using Core}

AuraID provides the following runtime capabilities:
\begin{itemize}
    \item \textbf{Workspace access}, including reading, creating, and modifying files within an experiment-specific project space;
    \item \textbf{Command-line execution}, enabling invocation of scientific scripts, local tools, package environments, and auxiliary utilities;
    \item \textbf{Multi-model compatibility}, allowing different large-model APIs to be connected according to performance, cost, privacy, or deployment constraints;
    \item \textbf{Autonomous agent loop}, enabling the system to decompose a user request into multi-step reasoning and action sequences.
\end{itemize}

Unlike developer-facing coding agents, however, AuraID intentionally deemphasizes direct code editing as the primary user interface. Instead, it adopts a \emph{conversation-first} interaction model. Users express objectives, constraints, and desired changes in natural language, while the agent performs the corresponding code synthesis, file modification, parameter revision, or tool execution in the background.

This design reflects the reality that experimental researchers are typically \emph{goal-oriented rather than code-oriented}. Their intent is to define analytical logic, software operation targets, or workflow constraints, not to participate in a manual edit-run-debug cycle. By placing natural-language interaction above a code-capable execution core, AuraID preserves flexibility while lowering the barrier for non-programmer users.

\subsubsection{Extensibility through Importable Skills}

A further requirement in characterization environments is heterogeneity. Different laboratories use different instruments, software stacks, analysis conventions, and reporting requirements. No closed set of built-in capabilities can adequately cover such diversity. AuraID therefore supports importing reusable skills from external repositories such as GitHub and ClawHub.

This design makes the platform incrementally extensible. Laboratories can assemble a customized capability set by adding analysis routines, GUI operating procedures, or workflow templates relevant to their own equipment and protocols. More importantly, this repository-based mechanism enables a transition from one-off success to reusable capability: once a useful behavior has been validated, it can be packaged as a skill and reused across future tasks rather than rediscovered from scratch.

\subsection{Skill-Centric Capability Accumulation}

\subsubsection{Why Skills Are Needed}

The practical bottleneck in scientific software automation is not merely action generation, but the acquisition, representation, and reuse of domain-specific procedural knowledge. This is especially evident in GUI-based instrument software.

Modern GUI agents and computer-use agents can in principle interact with arbitrary software interfaces. However, in our setting, we observed that their zero-shot performance on previously unseen scientific instrument software is often unreliable and far from deployment-ready. This limitation should not be understood simply as a failure of the underlying model. Human operators also require training before they can competently use complex characterization software. In real laboratories, novices learn through manuals, expert demonstration, supervised repetition, and gradual internalization of tacit heuristics. Instrument software often involves irregular UI design, weakly standardized interaction patterns, implicit state transitions, and domain-specific decision points. Under these conditions, expecting robust zero-shot operation is unrealistic.

A central question, then, is how the agent should \emph{store what it learns}. A straightforward idea is model fine-tuning. We do not adopt this as the primary adaptation mechanism because it is poorly matched to the deployment rhythm of laboratory environments. Fine-tuning GUI behavior typically requires large collections of trajectory data, specialized compute resources, hyperparameter tuning, and repeated retraining whenever new software or procedures must be incorporated. In contrast, scientific labs need a mechanism for rapidly onboarding new instruments, new procedures, and local operational variations with low latency and modest engineering effort.

Another possible strategy is to inject manuals, instructions, or external knowledge bases directly into prompts. While such contextual augmentation is useful for reference, we found it insufficient for executing long-horizon scientific procedures. Many instrument workflows span dozens of actions and contain a few difficult transitions that dominate failure. Purely textual guidance does not reliably solve problems of long-horizon drift, weak grounding between textual instructions and GUI state, or adaptive decision-making at critical steps.

These observations motivate our choice of \emph{skill} as the primary unit of accumulated capability. Inspired by skill-oriented agent frameworks in coding agents and by recent work showing that GUI agents can also benefit from structured skills, we treat a skill as a reusable, parameterizable, and composable unit of procedural knowledge. In this formulation, the system does not primarily grow through repeated model retraining, but through the accumulation and refinement of skills. A transient successful behavior becomes persistent system capability only after it is abstracted into an explicit skill artifact.

\subsubsection{Two Types of Skills}

According to their construction pathway and execution substrate, AuraID organizes skills into two categories: \textit{Type-1 GUI operational skills} and \textit{Type-2 analytical script skills}. This taxonomy reflects two distinct but complementary forms of expertise in scientific characterization.

Type-1 skills target \emph{instrument software operation}. Their knowledge source is expert demonstration over GUI interfaces, and their execution substrate is a computer-use agent loop interacting directly with software screens and controls. In implementation, they are instantiated as CUA-based GUI skills. Type-2 skills target \emph{post-acquisition data interpretation}. Their knowledge source is expert analytical intent expressed through natural language, and their execution substrate is a scriptable computational environment. In implementation, they are instantiated as conversationally constructed script skills.

Together, these two skill types cover the two major loci of capability required in characterization: operational control over proprietary software and digital scientific analysis.

\paragraph{Type-1 Skills: GUI Operational Skills.}

Type-1 skills are the mechanism through which AuraID addresses the software interaction barrier posed by proprietary scientific instrument systems. These skills are intended for GUI-based instrument software and are constructed through collaboration between human experts and the AuraID agent. Rather than assuming the existence of APIs, AuraID learns to operate instruments through the same interface layer used by human operators.

For complex procedures such as CT image reconstruction, EBSD grain-boundary identification setup, or SEM/EDS operating sequences, human experts first demonstrate the standard workflow inside the instrument software. During this process, AuraID records the interaction trajectory, including mouse actions, keyboard inputs, and the associated visual changes on the screen. The demonstrated procedure is then abstracted into a structured and parameterizable skill.

A key design requirement is that such skills must not be reduced to static coordinate replay. Instead, they are represented as reusable execution flows with explicit parameters, state checks, and conditional branches. For example, an SEM imaging skill may expose accelerating voltage, working distance, and magnification as configurable parameters, while preserving the underlying procedure for menu navigation, focus adjustment, scan initiation, and result export. This allows one operational logic to be instantiated across different samples and imaging needs.

More importantly, Type-1 skills can also encode tacit expert heuristics. In real instrument operation, important decisions are often made based on subtle cues that are difficult to formalize in APIs or manuals. During EDS acquisition, for instance, an experienced operator may prolong collection time when the live preview suggests insufficient signal-to-noise ratio, or terminate early once spectral quality becomes acceptable. Such behaviors can be represented inside the skill as conditional triggers, adaptive stopping criteria, or branching decisions tied to observed GUI state. As a result, Type-1 skills become not merely executable procedures, but compact carriers of operational expertise.

\paragraph{Type-2 Skills: Analytical Script Skills.}

Type-2 skills are designed for the digital analysis stage following physical characterization. In real scientific practice, laboratories often adopt different preferences for processing the same modality of data, such as UV-vis, FTIR, Raman, XRD, or electrochemical measurements. As a result, fixed commercial pipelines are often too rigid to satisfy the flexibility required in research-oriented analysis. AuraID addresses this problem through a conversational script-generation paradigm, allowing researchers to specify analytical logic in natural language and iteratively refine it through dialogue with the agent.

In a typical interaction, the user describes the intended procedure at an operational level, such as loading a CSV file, finding a peak in a target interval, performing baseline correction, applying Gaussian fitting, and exporting a figure and summary table. Guided by these instructions, the AuraID agent uses code generation together with domain-relevant scientific libraries such as NumPy, SciPy, pandas, matplotlib, or RDKit to synthesize executable analysis scripts. The scripts can be run immediately on experimental data, allowing the user to inspect intermediate results and visualizations and to iteratively refine the procedure.

Once validated, the script is abstracted into a standardized skill artifact with explicit input-output definitions, metadata, dependency specifications, and execution requirements. In this way, an ad hoc analytical interaction is converted into a reusable computational capability. Future tasks can invoke the same analysis logic directly rather than regenerating it from scratch.

The significance of Type-2 skills lies in their ability to bridge scientific expertise and executable analysis. Researchers do not need to manually program each routine, yet the resulting procedures remain transparent, customizable, and reusable. This is particularly valuable in characterization contexts where analysis standards differ across laboratories and evolve with research questions.

\subsubsection{Functional Relationship Between the Two Skill Types}

The two skill types are complementary rather than isolated. In a typical characterization workflow, a Type-1 skill may operate instrument software to configure acquisition parameters, launch measurement, reconstruct results, and export data. The exported data can then be passed to a Type-2 skill for peak fitting, quantitative analysis, feature extraction, or report generation. The analytical outcome may further inform the next round of operational decisions, producing a closed-loop workflow that links physical experimentation, software interaction, and digital interpretation.

This complementarity is central to the practical value of AuraID. Type-1 skills address the absence of open interfaces in real-world instrument ecosystems, while Type-2 skills address variability and customization in scientific analysis. Their combination forms the reusable knowledge substrate on which AuraID's intelligent behavior depends.

\begin{figure}[t]
    \centering
    \includegraphics[width=\textwidth]{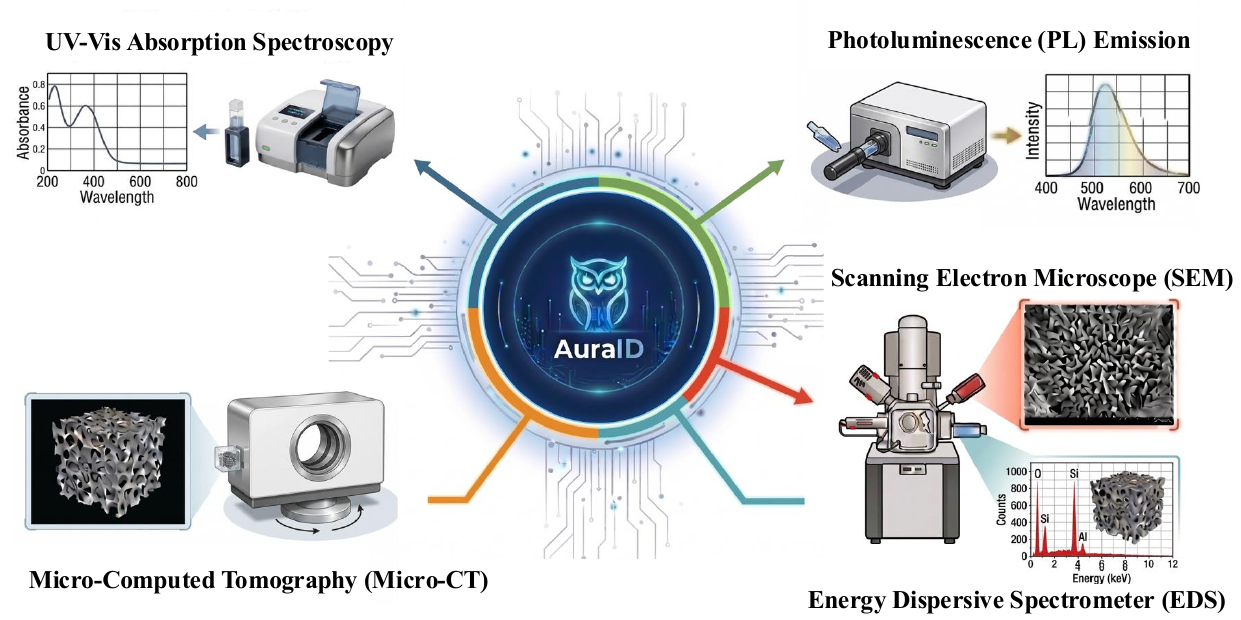} 
    \caption{The platform integrates fluorescence emission spectroscopy with UV-Vis absorption spectroscopy, SEM, EDS, and Micro-CT for multi-modal characterization.}
    \label{fig:instrument-list}
\end{figure}

\section{Intelligent Driving of Scientific Instruments}\label{sec4}

AuraID has been developed toward intelligent automation across a broad range of scientific characterization instruments, covering five categories of precision equipment (\Cref{fig:instrument-list}). This section presents several representative scenarios to illustrate how embodied manipulation, GUI-based software operation, and analytical skill execution are integrated into end-to-end characterization workflows.

\subsection{Intelligent Spectral Characterization}

Spectral characterization is one of the most common and foundational procedures in chemistry and materials science. AuraID supports a closed-loop workflow for multimodal spectral analysis, including sample handling, instrument operation, and automated interpretation of acquired spectra.

\begin{figure}[t]
    \centering
    \includegraphics[width=\linewidth]{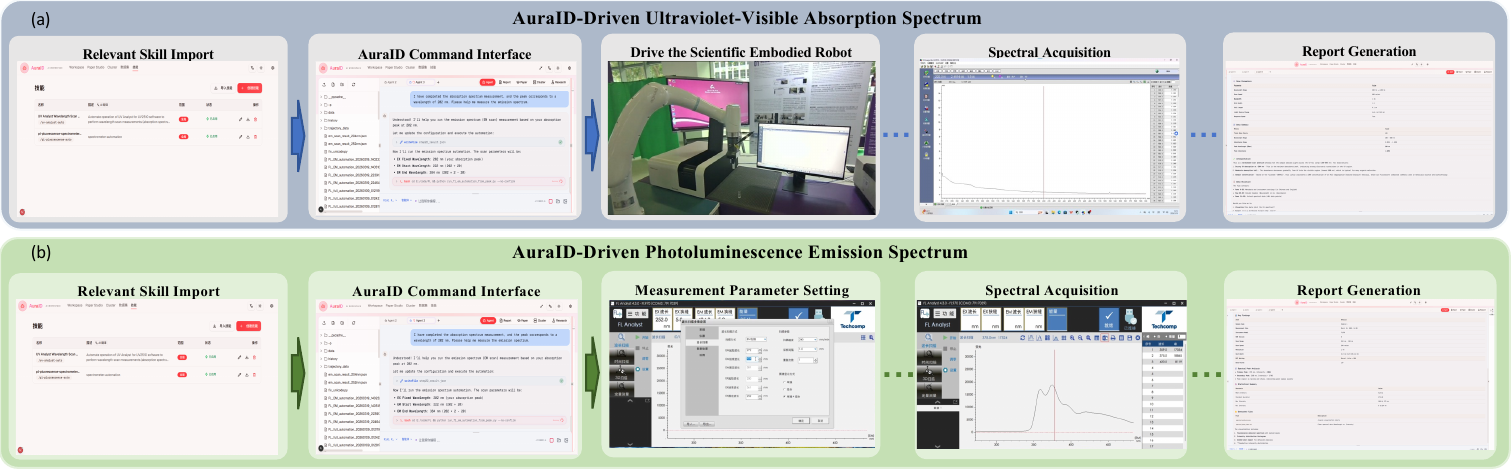}
    \caption{AuraID workflow for spectral characterization of UV-vis and PL.}
    \label{fig:uvvis_pl_pipeline}
\end{figure}

\paragraph{UV-Vis Absorption Spectroscopy (UV-vis) and Photoluminescence (PL) Emission}
UV-vis and PL spectroscopy provide essential information about the electronic band structures and optical transition behaviors of functional materials.
For UV-vis and PL measurements (\Cref{fig:uvvis_pl_pipeline}), AuraID initiates with the embodied agent executing high-precision sample handling and exchange. Utilizing multi-modal perception, the agent ensures sub-millimeter alignment of cuvettes or solid substrates within the sample chamber, maintaining rigorous physical consistency across experimental batches.

Once the physical setup is secured, the system deploys CUA-driven operational skills to interface directly with the instrument’s native graphical software. These agents mimic human cognitive patterns to navigate complex UI hierarchies, autonomously configuring essential acquisition parameters. By interpreting real-time visual feedback from the software, the CUA-based skills can proactively handle operational exceptions and dynamically optimize measurement settings to ensure the acquisition of optimal characterization results.

\subsection{Intelligent Microscopic Imaging Characterization}

Microscopic imaging presents a unique challenge for laboratory autonomy, as achieving high-fidelity results traditionally necessitates continuous, expert-level adjustments based on real-time visual feedback. AuraID transcends static automation by enabling adaptive, closed-loop control across complex imaging modalities, including Scanning Electron Microscopy (SEM), Energy-Dispersive Spectroscopy (EDS), and Micro-Computed Tomography (Micro-CT).

\begin{figure}[t]
    \centering
    \includegraphics[width=\linewidth]{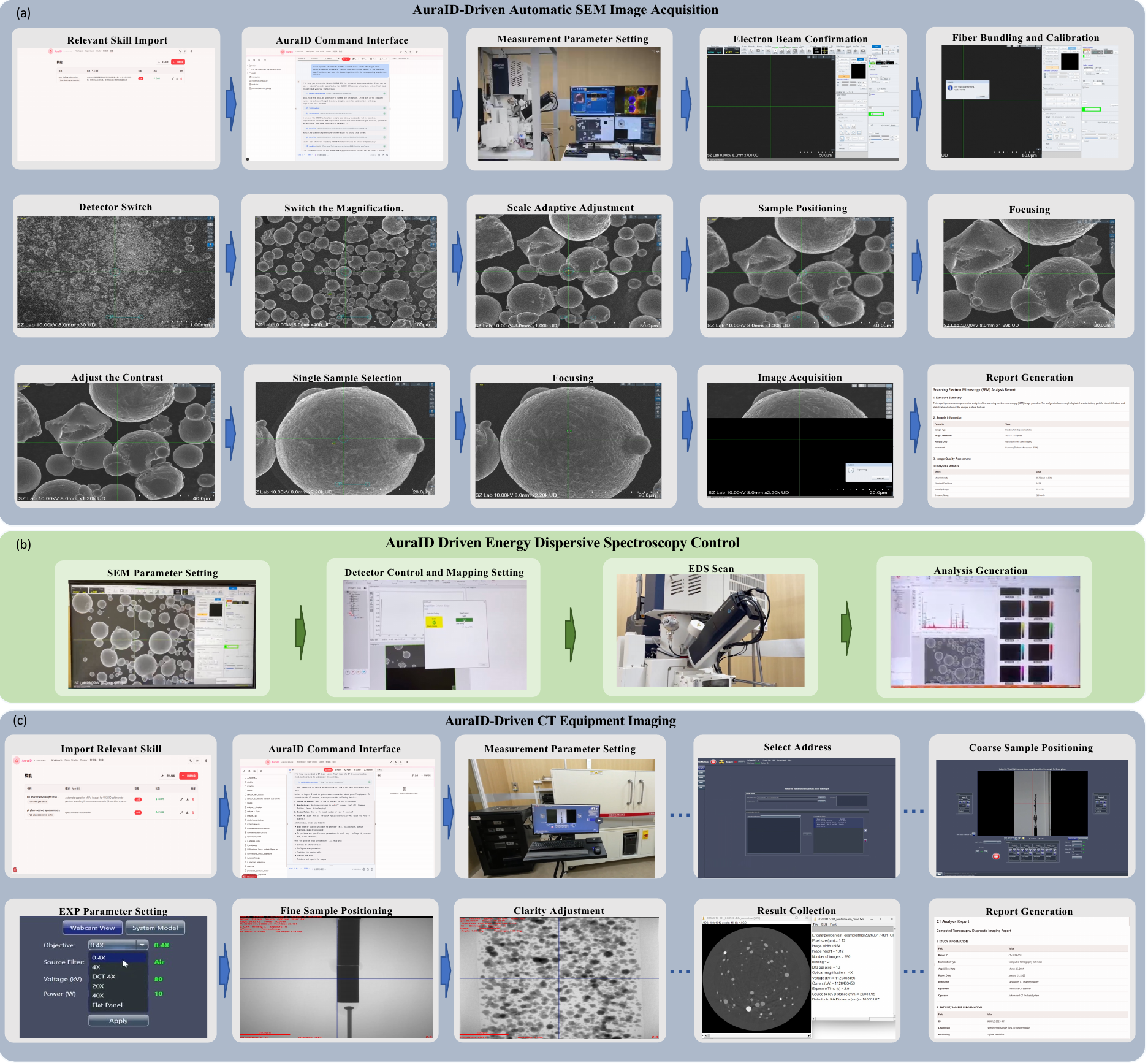}
    \caption{AuraID workflow for microscopic imaging characterization of SEM, Micro-CT, and EDS.}
    \label{fig:sem_ct_pipeline}
\end{figure}

\paragraph{Scanning Electron Microscopy}
Scanning electron microscopy provides high-resolution topographic and compositional information essential for characterizing material surfaces and microstructures.
For these experiments, as shown in~\Cref{fig:sem_ct_pipeline}a, AuraID manages a complete loop from sample loading to high-fidelity imaging. The process begins with the embodied agent using a robotic arm to place samples into the vacuum chamber with precision. Once the chamber is sealed, the system uses CUA-driven skills to operate the control software, autonomously initiating vacuum sequences and performing sample localization. During imaging, AuraID uses real-time visual feedback to adjust critical parameters such as accelerating voltage, magnification, and focus. This closed-loop approach mimics the adjustments of an expert operator to ensure optimal contrast and resolution, preparing the system for subsequent quantitative analysis.

\paragraph{Energy-Dispersive Spectroscopy}
Energy-dispersive spectroscopy provides qualitative and quantitative identification of elemental compositions and their spatial distributions within a sample. As an integrated accessory to the SEM, EDS enables AuraID to perform in-situ chemical analysis following morphological imaging (\Cref{fig:sem_ct_pipeline}b). The system operates the control software to identify target regions of interest (ROIs) and configure excitation parameters, such as accelerating voltage and process time, for optimal X-ray signal collection.

\paragraph{Micro-Computed Tomography}
Micro-computed tomography visualizes and quantifies the three-dimensional internal structures and density distributions of physical samples. 
For these measurements (\Cref{fig:sem_ct_pipeline}c), AuraID interfaces with the instrument's control system to ensure stable sample positioning throughout the multi-angle projection process. Based on the experimental objective, the agent autonomously configures scanning parameters, including spatial resolution and acquisition duration, to optimize the signal-to-noise ratio. The system manages the entire data pipeline, from raw projection collection to the execution of tomographic reconstruction software. This integrated workflow enables the automated derivation of detailed internal morphology and volumetric information, effectively creating a digital twin of the sample's interior.

Taken together, these applications demonstrate that AuraID extends beyond isolated software automation or device-specific scripting. Instead, it establishes a unified, skill-based framework that seamlessly integrates physical manipulation, GUI-native interaction, and scientific data acquisition across heterogeneous characterization instruments.
This versatile architecture enables the system to adapt to diverse experimental requirements, transforming heterogeneous instrumentation into a unified, autonomous characterization ecosystem.

\section{Intelligent Analysis of Scientific Data}\label{sec5}

Beyond intelligent driving of scientific instruments, AuraID further demonstrates its capacity for the intelligent analysis of scientific data. The system autonomously executes advanced data interpretation tasks.

\subsection{Spectral Analysis}

\begin{figure}[t]
    \centering
    \includegraphics[width=\linewidth]{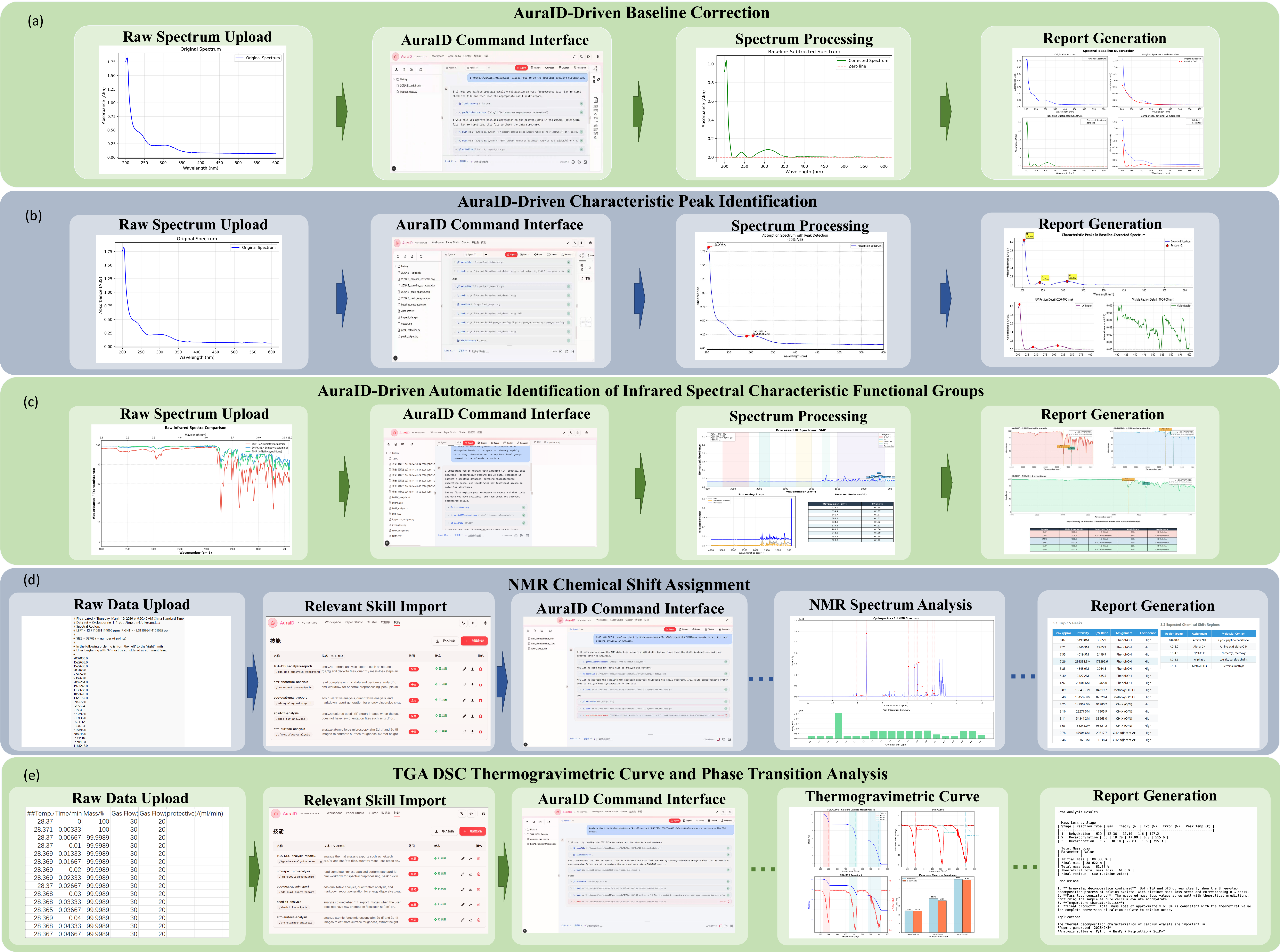}
    \caption{AuraID workflow for spectral analysis.}
    \label{fig:spectral_analysis_pipeline}
\end{figure}

\paragraph{Baseline Correction}

For raw UV-vis data, AuraID intelligently identifies and eliminates background noise originating from instrumental fluctuations and stray light scattering. This automated baseline correction (\Cref{fig:spectral_analysis_pipeline}a) is critical for removing non-linear offsets, ensuring that extracted peak attributes accurately represent the intrinsic electronic properties of the sample.
Furthermore, this analytical framework can be extended to validate additional spectroscopic data, such as X-ray Diffraction (XRD) and Raman spectroscopy, in future studies.

\paragraph{Peak Identification}

Based on the baseline-corrected spectra, AuraID intelligently identifies the specific positions and intensities of absorption peaks (\Cref{fig:spectral_analysis_pipeline}b). By automatically extracting these key peak parameters, the system eliminates the labor-intensive process of manual inspection and comparative analysis. 
This functionality is easily extendable to peak identification across other spectroscopic modalities.

\paragraph{Functional Group Assignment in Fourier Transform Infrared (FTIR) Spectroscopy}

Fourier transform infrared spectroscopy is a tool for identifying molecular structures and chemical bonds by probing the vibrational modes of functional groups. 
AuraID processes raw FTIR spectral data by autonomously cross-referencing expansive infrared databases to achieve precision matching of characteristic absorption bands (\Cref{fig:spectral_analysis_pipeline}c). 
This capability enables the rapid identification and output of key functional group compositions within complex molecular frameworks. 
The challenge of FTIR interpretation often arises from the overlapping of vibrational modes and the influence of hydrogen bonding, which can shift absorption frequencies. 
By digitizing expert recognition logic, AuraID effectively deciphers these spectral fingerprints, transforming raw vibrational signals into definitive structural information with expert-level accuracy.

\paragraph{Nuclear Magnetic Resonance (NMR) Spectrum Interpretation}

Nuclear magnetic resonance spectroscopy provides high-resolution insights into the local magnetic environments of atomic nuclei, serving as a primary tool for definitive molecular structure elucidation. 
For raw NMR data, AuraID autonomously executes essential signal processing sequences, including phase correction, baseline calibration, and peak area integration (\Cref{fig:spectral_analysis_pipeline}d). 
By integrating with chemical shift databases, the system identifies resonance frequencies and determines the relative stoichiometry of protons or other nuclei. 
By automating these intricate adjustments, AuraID ensures a high-fidelity transition from raw electromagnetic induction signals to precise structural assignments, significantly reducing the manual effort required for complex spectra.

\paragraph{Thermal Analysis with Thermogravimetric analysis (TGA)}

Thermogravimetric analysis is a fundamental technique for assessing the thermal stability and compositional shifts of materials by monitoring mass change as a function of temperature. 
AuraID autonomously identifies critical weight-loss steps and thermal reaction peaks within the raw thermometric curves (\Cref{fig:spectral_analysis_pipeline}e). 
By applying automated tangential and integration methods, the system precisely calculates essential thermodynamic parameters, including decomposition temperatures ($T_d$) and reaction enthalpies ($\Delta H$). 
By digitizing these analytical workflows, AuraID transforms complex raw thermal profiles into definitive kinetic and thermodynamic insights with high consistency.

\subsection{Microscopic Imaging Analysis}

\begin{figure}[t]
    \centering
    \includegraphics[width=\linewidth]{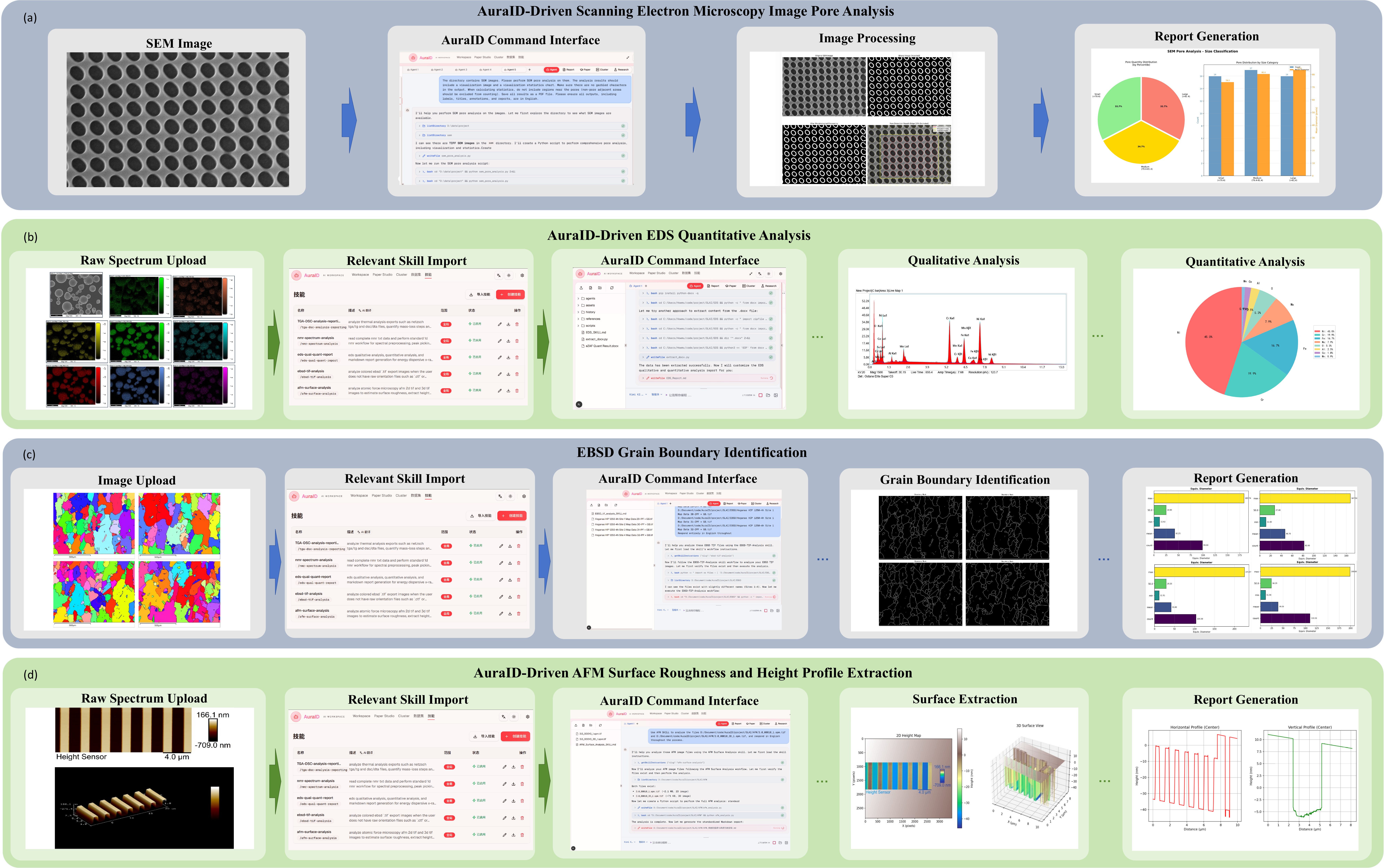}
    \caption{AuraID workflow for microscopic imaging analysis.}
    \label{fig:micro_image_analysis_pipeline}
\end{figure}

\paragraph{SEM Analysis}

Quantitative analysis of scanning electron microscopy images is essential for correlating material morphology, such as porosity and particle distribution, with macroscopic performance. 
AuraID autonomously detects and delineates pore contours within high-resolution SEM micrographs (\Cref{fig:micro_image_analysis_pipeline}a). 
By executing high-throughput batch-processing, the system performs large-scale quantification and generates objective statistical distributions of structural parameters.
The primary challenge of manual interpretation lies in the labor-intensive nature of manual quantification, which is often confounded by low throughput and subjective bias arising from variations in image contrast and feature overlapping. 
AuraID turns expert manual inspection into automated algorithms, making it easy to get consistent measurement results from visual data.
Looking forward, the extension of this framework to in-situ SEM analysis will enable the real-time monitoring of dynamic material evolution, such as structural deformation or phase transformation, under external stimuli.

\paragraph{EDS Analysis}
Energy-dispersive spectroscopy provides essential quantification of elemental compositions and their spatial distributions within a sample. 
By integrating the scanning results, the system autonomously identifies elemental species and resolves spectral interference from overlapping peaks (\Cref{fig:micro_image_analysis_pipeline}b). 
AuraID then calculates the weight and atomic percentages for each element and exports standardized compositional analysis reports. 
By automating these analytical procedures, AuraID transforms raw X-ray counts into precise chemical information, ensuring high consistency and efficiency in multi-element characterization.

\paragraph{Atomic Force Microscopy (AFM) Analysis}
Atomic Force Microscopy is a high-resolution imaging technique used to characterize the three-dimensional surface topography and physical properties of materials at the nanoscale. 
By processing raw AFM scanning data, AuraID autonomously calculates surface roughness parameters, such as arithmetic average ($Ra$) and root mean square ($Rq$) (\Cref{fig:micro_image_analysis_pipeline}c). 
The system precisely extracts height profile curves from designated regions of interest to quantify microscopic surface features and fluctuations. 
The manual analysis of AFM data is often complicated by the need for meticulous image leveling and tip-convolution corrections to ensure accurate vertical resolution. 
By automating these analytical steps, AuraID converts raw topographic maps into standardized morphological metrics, ensuring high consistency in thin-film and surface characterization.

\paragraph{Electron Backscatter Diffraction (EBSD) Analysis}
Electron backscatter diffraction is a scanning electron microscopy technique used to characterize the crystallographic orientation and microstructural evolution of crystalline materials. 
By processing raw EBSD data, AuraID autonomously generates inverse pole figure (IPF) maps and identifies various types of grain boundaries (\Cref{fig:micro_image_analysis_pipeline}d). 
The system precisely extracts statistical distributions of grain sizes and misorientation angles to quantify the underlying microstructural architecture. 
This automation addresses the primary bottlenecks in manual EBSD interpretation, specifically the meticulous indexing of diffraction patterns and the robust segmentation of fine-grained structures amidst background noise.



\section{Discussion}\label{sec6}

In this work, we show that autonomous scientific characterization in heterogeneous real-world laboratories can be formulated as a GUI-native computer-use problem, rather than only as an API- or CLI-mediated automation task. By combining a conversation-first agent runtime with reusable Type-1 GUI operational skills and Type-2 analytical script skills, AuraID connects physical sample handling, proprietary instrument software operation, workspace computation, and scientific data interpretation within a unified characterization workflow. These results suggest that scientific software itself can serve as a practical operating environment for laboratory intelligence, particularly in settings where vendor-specific interfaces, closed-source control stacks, and visually grounded decision-making remain central to routine practice.

AuraID owns a distinct position within the landscape of autonomous research by extending laboratory intelligence into the underexplored domain of GUI-centric scientific characterization. 
While conventional self-driving laboratories often rely on pre-engineered APIs and fixed automation stacks—limiting their scope to standardized synthesis or liquid-handling—AuraID’s GUI-native paradigm bypasses the requirement for stable programmatic endpoints. 
By directly navigating the proprietary interfaces and closed-source control stacks inherent to modern instrumentation, this framework enables the rapid unification of diverse modalities into a single, cohesive analytical continuum. 
This multi-instrument integration allows for the simultaneous interrogation of structural, elemental, and functional properties. 
Ultimately, by capturing the multi-dimensional descriptors required for complex materials challenges, AuraID transforms heterogeneous characterization workflows from isolated manual tasks into a holistic, automated engine for scientific discovery.

Beyond automation, AuraID serves as a critical engine for high-quality scientific data acquisition. The ``reusable skill'' architecture ensures that experimental procedures are executed with expert-level precision. 
This consistency is important because, by systematically converting manual laboratory procedures into structured, machine-readable datasets, we provide the essential high-quality scientific data for the next generation of Scientific Foundation Models. 
As these models evolve, the high-fidelity data generated through GUI-native agents will allow AI to not only execute experiments but to develop a deeper, cross-domain understanding of experimental physics and chemistry.

While AuraID demonstrates robust coverage across diverse modalities, several frontiers remain. 
The current reliance on expert demonstration for skill acquisition suggests a need for self-supervised skill induction, where agents learn to navigate new software via curiosity-driven exploration or by ``reading'' technical manuals. 
Furthermore, while we have bridged the software gap, the physical handling of sensitive samples across instruments still requires sophisticated robotic coordination.

Looking forward, the convergence of GUI-native autonomy and multi-modal characterization points toward a continually evolving laboratory intelligence. 
By capturing and digitizing the tacit knowledge of human experts, we are building a persistent, composable capability that transcends individual experiments. 
As we integrate more instruments and refine the quality of the resulting data, these agents will eventually move from being mere operators to becoming autonomous co-scientists, capable of closing the loop between hypothesis generation and complex physical validation. 
This trajectory offers a practical and scalable path toward the fully autonomous, AI-driven laboratories of the future.

\backmatter

\bibliography{bib/sn-bibliography, bib/1-intro.bib, bib/2-related-work.bib}

@article{AlphaFold2021,
  author = {Jumper, John and Evans, Richard and Pritzel, Alexander and Green, Tim and Figurnov, Michael and Ronneberger, Olaf and Tunyasuvunakool, Kathryn and Bates, Russ and Zidek, Augustin and Potapenko, Anna and Bridgland, Alex and Meyer, Clemens and Kohl, Simon A A and Ballard, Andrew J and Cowie, Andrew and Romera-Paredes, Bernardino and Nikolov, Stanislav and Jain, Rishub and Adler, Jonas and Back, Trevor and Petersen, Stig and Reiman, David and Clancy, Ellen and Zielinski, Michal and Steinegger, Martin and Pacholska, Michalina and Berghammer, Tamas and Bodenstein, Sebastian and Silver, David and Vinyals, Oriol and Senior, Andrew W and Kavukcuoglu, Koray and Kohli, Pushmeet and Hassabis, Demis},
  title = {Highly accurate protein structure prediction with AlphaFold},
  journal = {Nature},
  year = {2021},
  volume = {596},
  number = {7873},
  pages = {583--589},
  doi = {10.1038/s41586-021-03819-2}
}

@article{AlphaFold3_2024,
  author = {Abramson, Josh and Adler, Jonas and Dunger, Jack and Evans, Richard and Green, Tim and Pritzel, Alexander and Ronneberger, Olaf and Willmore, Lindsay and Ballard, Andrew J and Bambrick, Joshua},
  title = {Accurate structure prediction of biomolecular interactions with AlphaFold 3},
  journal = {Nature},
  year = {2024},
  volume = {630},
  number = {8016},
  pages = {493--500},
  doi = {10.1038/s41586-024-07487-w}
}

@article{GNoME2023,
  author = {Merchant, Amil and Batzner, Simon and Schoenholz, Samuel S. and Aykol, Muratahan and Cheon, Gowoon and Cubuk, Ekin Dogus},
  title = {Scaling deep learning for materials discovery},
  journal = {Nature},
  year = {2023},
  volume = {624},
  pages = {80--85},
  doi = {10.1038/s41586-023-06735-9}
}

@article{ALab2023,
  author = {Szymanski, Nathan J. and Rendy, Bernardus and Fei, Yuxing and Kumar, Rishi E. and He, Tanjin and Milsted, David and McDermott, Matthew J. and Gallant, Max and Cubuk, Ekin Dogus and Merchant, Amil and Kim, Haegyeom and Jain, Anubhav and Bartel, Christopher J. and Persson, Kristin and Zeng, Yan and Ceder, Gerbrand},
  title = {An autonomous laboratory for the accelerated synthesis of novel materials},
  journal = {Nature},
  year = {2023},
  volume = {624},
  number = {7990},
  pages = {86--91},
  doi = {10.1038/s41586-023-06734-w}
}

@article{MobileRoboticChemist2020,
  author = {Burger, Benjamin and Maffettone, Phillip M. and Gusev, Vladimir V. and Aitchison, Catherine M. and Bai, Yang and Wang, Xiaoyan and Li, Xiaobo and Alston, Ben M. and Li, Buyi and Clowes, Rob and Rankin, Nicola and Harris, Brandon and Sprick, Reiner Sebastian and Cooper, Andrew I},
  title = {A mobile robotic chemist},
  journal = {Nature},
  year = {2020},
  volume = {583},
  number = {7815},
  pages = {237--241},
  doi = {10.1038/s41586-020-2442-2}
}

@article{OSWorld2024,
  author = {Xie, Tianbao and Zhang, Danyang and Chen, Jixuan and Li, Xiaochuan and Zhao, Siheng and Cao, Ruisheng and Hua, Toh Jing and Cheng, Zhoujun and Shin, Dongchan and Lei, Fangyu and Liu, Yitao and Xu, Yiheng and Zhou, Shuyan and Savarese, Silvio and Xiong, Caiming and Zhong, Victor and Yu, Tao},
  title = {OSWorld: Benchmarking Multimodal Agents for Open-Ended Tasks in Real Computer Environments},
  journal = {arXiv preprint arXiv:2404.07972},
  year = {2024},
  eprint = {2404.07972},
  url = {https://arxiv.org/abs/2404.07972}
}

@article{AndroidWorld2024,
  author = {Rawles, Christopher and Clinckemaillie, Sarah and Chang, Yifan and Waltz, Jonathan and Lau, Gabrielle and Fair, Marybeth and Li, Alice and Bishop, William and Li, Wei and Campbell-Ajala, Folawiyo and Toyama, Daniel and Berry, Robert and Tyamagundlu, Divya and Lillicrap, Timothy and Riva, Oriana},
  title = {{AndroidWorld}: A Dynamic Benchmarking Environment for Autonomous Agents},
  journal = {arXiv preprint arXiv:2405.14573},
  year = {2024},
  eprint = {2405.14573},
  doi = {10.48550/arXiv.2405.14573},
  url = {https://arxiv.org/abs/2405.14573}
}

@article{UITARS2025,
  author = {Qin, Yujia and Ye, Yining and Fang, Junjie and Wang, Haoming and Liang, Shihao and Tian, Shizuo and Zhang, Junda and Li, Jiahao and Li, Yunxin and Huang, Shijue and Zhong, Wanjun and Li, Kuanye and Yang, Jiale and Miao, Yu and Lin, Woyu and Liu, Longxiang and Jiang, Xu and Ma, Qianli and Li, Jingyu and Xiao, Xiaojun and Cai, Kai and Li, Chuang and Zheng, Yaowei and Jin, Chaolin and Li, Chen and Zhou, Xiao and Wang, Minchao and Chen, Haoli and Li, Zhaojian and Yang, Haihua and Liu, Haifeng and Lin, Feng and Peng, Tao and Liu, Xin and Shi, Guang},
  title = {{UI-TARS}: Pioneering Automated {GUI} Interaction with Native Agents},
  journal = {arXiv preprint arXiv:2501.12326},
  year = {2025},
  eprint = {2501.12326},
  doi = {10.48550/arXiv.2501.12326},
  url = {https://arxiv.org/abs/2501.12326}
}

@article{team2026kimi,
  title = {Kimi {K2.5}: Visual Agentic Intelligence},
  author = {Team, Kimi and Bai, Tongtong and Bai, Yifan and Bao, Yiping and Cai, SH and Cao, Yuan and Charles, Y and Che, HS and Chen, Cheng and Chen, Guanduo and others},
  journal = {arXiv preprint arXiv:2602.02276},
  year = {2026},
  eprint = {2602.02276},
  url = {https://arxiv.org/abs/2602.02276},
  doi = {10.48550/arXiv.2602.02276}
}

@misc{CLIAnything2025,
  author = {{HKUDS}},
  title = {{CLI-Anything}: Making {ALL} Software Agent-Native},
  year = {2025},
  howpublished = {GitHub repository},
  url = {https://github.com/HKUDS/CLI-Anything},
  note = {Software project (not a traditional article). License and citation: see repository \texttt{README.md}.}
}

@misc{OpenCLI2025,
  author = {{jackwener}},
  title = {{OpenCLI}: Make Any Website \& Tool Your {CLI}},
  year = {2025},
  howpublished = {GitHub repository; npm package \texttt{@jackwener/opencli}},
  url = {https://github.com/jackwener/opencli},
  note = {Software project (not a traditional article).}
}

@misc{claude4_6,
  author = {{Anthropic}},
  title = {Claude {Sonnet} 4.6},
  year = {2026},
  howpublished = {Research / product documentation},
  url = {https://www.anthropic.com/research/claude-sonnet-4-6},
  note = {Official Anthropic page for Sonnet 4.6 capabilities and availability.}
}

@misc{gpt54,
  author = {{OpenAI}},
  title = {Introducing {GPT-5.4}},
  year = {2026},
  howpublished = {Company announcement},
  url = {https://openai.com/index/introducing-gpt-5-4/},
  note = {Official OpenAI announcement for GPT-5.4 (including computer-use and benchmark discussion).}
}

@misc{seed1_8,
  author = {{ByteDance Seed Team}},
  title = {Official Release of {Seed} 1.8: A Generalized Agentic Model},
  year = {2025},
  howpublished = {Official blog post},
  url = {https://seed.bytedance.com/en/blog/official-release-of-seed1-8-a-generalized-agentic-model},
  note = {See also repository \texttt{ByteDance-Seed/Seed-1.8} on GitHub for model card and artifacts.}
}

@article{SDLReview2025,
  author = {Tobias, Alexander V. and Wahab, Adam},
  title = {Autonomous {`}self-driving{' laboratories}: a review of technology and policy implications},
  journal = {R. Soc. Open Sci.},
  year = {2025},
  volume = {12},
  number = {7},
  pages = {250646},
  doi = {10.1098/rsos.250646}
}

@article{AIDrivenAutonomousLab2025,
  author = {Chen, Junwu and Xu, Qiucheng},
  title = {Artificial intelligence-driven autonomous laboratory for accelerating chemical discovery},
  journal = {Chem. Synth.},
  year = {2025},
  volume = {5},
  pages = {76},
  doi = {10.20517/cs.2025.66}
}

@misc{AutonomousLabNovelMaterials,
  author = {{Lawrence Berkeley National Laboratory}},
  title = {Meet the {Autonomous Lab} of the Future},
  year = {2023},
  month = apr,
  howpublished = {Berkeley Lab News Center},
  url = {https://newscenter.lbl.gov/2023/04/17/meet-the-autonomous-lab-of-the-future/},
  note = {News article on the {A-Lab} autonomous synthesis facility at {LBNL}.}
}

@misc{CederAIInAction2026,
  author = {Ceder, Gerbrand},
  title = {Gerbrand Ceder: {AI} in action --- Autonomous laboratories for materials synthesis},
  howpublished = {YouTube video},
  organization = {{Bakar Institute of Digital Materials for the Planet}},
  address = {University of California, Berkeley},
  year = {2026},
  url = {https://youtu.be/gTZgbAgQkqc},
  note = {Seminar recording associated with the Bakar Institute for Digital Materials for the Planet (UC Berkeley) seminar series.}
}

@article{MobileRoboticProcessChemist,
  author = {Brass, Emma J. and Veeramani, Satheeshkumar and Zhou, Zhengxue and Fakhruldeen, Hatem and Manzano, J. Sebastian and Clowes, Rob and Akpinar, Isil and Ward, Miriam R. and Ward, John W. and Cooper, Andrew I.},
  title = {A mobile robotic process chemist},
  journal = {Digit. Discov.},
  year = {2026},
  volume = {5},
  number = {3},
  pages = {1363--1371},
  doi = {10.1039/D5DD00563A}
}

@article{AlabOS2024,
  author = {Fei, Yuxing and Rendy, Bernardus and Kumar, Rishi and Dartsi, Olympia and Sahasrabuddhe, Hrushikesh P. and McDermott, Matthew J. and Wang, Zheren and Szymanski, Nathan J. and Walters, Lauren N. and Milsted, David and Zeng, Yan and Jain, Anubhav and Ceder, Gerbrand},
  title = {{AlabOS}: A Python-based Reconfigurable Workflow Management Framework for Autonomous Laboratories},
  journal = {Digit. Discov.},
  year = {2024},
  volume = {3},
  number = {11},
  pages = {2275--2288},
  doi = {10.1039/D4DD00129J},
  note = {Preprint: arXiv:2405.13930.}
}

@article{Coscientist2023,
  author = {Boiko, Daniil A. and MacKnight, Robert and Kline, Ben and Gomes, Gabe},
  title = {Autonomous chemical research with large language models},
  journal = {Nature},
  year = {2023},
  volume = {624},
  pages = {570--578},
  doi = {10.1038/s41586-023-06792-0}
}

@article{ChemCrow2024,
  author = {Bran, Andres M. and Cox, Sam and Schilter, Oliver and Baldassari, Carlo and White, Andrew D. and Schwaller, Philippe},
  title = {Augmenting large language models with chemistry tools},
  journal = {Nat. Mach. Intell.},
  year = {2024},
  volume = {6},
  pages = {525--535},
  doi = {10.1038/s42256-024-00832-8}
}

@article{UniLabOS2025,
  author = {Gao, Jing and Chang, Junhan and Que, Haohui and Xiong, Yanfei and Zhang, Shixiang and Qi, Xianwei and Liu, Zhen and Wang, Jun-Jie and Ding, Qianjun and Li, Xinyu and Pan, Ziwei and Xie, Qiming and Yan, Zhuang and Yan, Junchi and Zhang, Linfeng},
  title = {{UniLabOS}: An {AI}-Native Operating System for Autonomous Laboratories},
  journal = {arXiv preprint arXiv:2512.21766},
  year = {2025},
  eprint = {2512.21766},
  url = {https://arxiv.org/abs/2512.21766}
}

@article{LabOS2025,
  author = {Cong, Le and Smerkous, David and Wang, Xiaotong and Yin, Di and Zhang, Zaixi and Jin, Ruofan and Wang, Yinkai and Gerasimiuk, Michal and Dinesh, Ravi K. and Smerkous, Alex and Shi, Lihan and Zheng, Joy and Lam, Ian and Wu, Xuekun and Liu, Shilong and Li, Peishan and Zhu, Yi and Zhao, Ning and Parakh, Meenal and Serrao, Simran and Mohammad, Imran A. and Chen, Chao-Yeh and Xie, Xiufeng and Chen, Tiffany and Weinstein, David and Barbone, Greg and Caglar, Belgin and Sunwoo, John B. and Li, Fuxin and Deng, Jia and Wu, Joseph C. and Wu, Sanfeng and Wang, Mengdi},
  title = {{LabOS}: The {AI}-{XR} Co-Scientist That Sees and Works With Humans},
  journal = {bioRxiv},
  year = {2025},
  doi = {10.1101/2025.10.16.679418},
  url = {https://www.biorxiv.org/content/10.1101/2025.10.16.679418}
}

@article{STELLA2025,
  author = {Jin, Ruofan and Xu, Mingyang and Meng, Fei and Wan, Guancheng and Cai, Qingran and Jiang, Yize and Han, Jin and Chen, Yuanyuan and Lu, Wanqing and Wang, Mengyang and Lan, Zhiqian and Jiang, Yuxuan and Liu, Junhong and Wang, Dongyao and Cong, Le and Zhang, Zaixi},
  title = {{STELLA}: Towards a Biomedical World Model with Self-Evolving Multimodal Agents},
  journal = {bioRxiv},
  year = {2025},
  doi = {10.1101/2025.07.01.662467},
  url = {https://www.biorxiv.org/content/10.1101/2025.07.01.662467}
}

@article{lu2026towards,
  title={Towards end-to-end automation of AI research},
  author={Lu, Chris and Lu, Cong and Lange, Robert Tjarko and Yamada, Yutaro and Hu, Shengran and Foerster, Jakob and Ha, David and Clune, Jeff},
  journal={Nature},
  volume={651},
  number={8107},
  pages={914--919},
  year={2026},
  publisher={Nature Publishing Group UK London}
}

@inproceedings{ToolingOrNotChemistryAgents,
  title = {Tooling or Not Tooling? The Impact of Tools on Language Agents for Chemistry Problem Solving},
  author = {Yu, Botao and Baker, Frazier N. and Chen, Ziru and Herb, Garrett and Gou, Boyu and Adu-Ampratwum, Daniel and Ning, Xia and Sun, Huan},
  booktitle = {Findings of the Association for Computational Linguistics: NAACL 2025},
  year = {2025},
  pages = {7620--7640},
  doi = {10.18653/v1/2025.findings-naacl.424},
  url = {https://aclanthology.org/2025.findings-naacl.424/},
  note = {Preprint: arXiv:2411.07228.}
}

@misc{InternSpectScientificDiscoveryPlatform,
  author = {{Shanghai Artificial Intelligence Laboratory}},
  title = {{InternSpect}},
  year = {2026},
  howpublished = {Project homepage},
  url = {https://internspect.intern-ai.org.cn/home},
  note = {Cited as a representative scientific data analysis platform.}
}

@misc{VenusSeries,
  author = {{Newtonoptic Research Institute}},
  title = {Venus Fully Automatic High-Throughput Cell Counter (Venus series)},
  year = {2026},
  howpublished = {Product page},
  url = {https://www.newtonoptic.com/products_details/2.html},
  note = {Representative Venus series instrument page by Newtonoptic.}
}

@misc{LGChemistryAutonomousLab,
  author = {{LG Chem}},
  title = {Autonomous Smart Lab (robot automation laboratory) at LG Chem Analytical Research Center},
  year = {2025},
  month = sep,
  howpublished = {Press release},
  url = {https://www.lg.co.kr/media/release/29390},
  note = {Official LG press release announcing LG Chem's Autonomous Smart Lab (robot automation laboratory).}
}

@misc{ChemspeedPlatform,
  author = {{Chemspeed Technologies AG}},
  title = {Chemspeed: Automation and Digitalization of {R\&D} and {QC} Labs},
  year = {2026},
  howpublished = {Company homepage},
  url = {https://www.chemspeed.com/},
  note = {Cited as a representative commercial lab automation platform provider.}
}

@article{LUMILab,
  author = {Cui, Haotian and Xu, Yue and Pang, Kuan and Li, Gen and Gong, Fanglin and Wang, Bo and Li, Bowen},
  title = {{LUMI}-lab: a Foundation Model-Driven Autonomous Platform Enabling Discovery of New Ionizable Lipid Designs for mRNA Delivery},
  journal = {bioRxiv},
  year = {2025},
  doi = {10.1101/2025.02.14.638383},
  url = {https://www.biorxiv.org/content/10.1101/2025.02.14.638383}
}

@misc{InnoClaw2026,
  author = {{SpectrAI Initiative}},
  title = {{InnoClaw}: A self-hostable {AI} research workspace for grounded chat, paper study, scientific skills, and research execution},
  year = {2026},
  howpublished = {GitHub repository},
  url = {https://github.com/SpectrAI-Initiative/InnoClaw},
  note = {Apache-2.0 licensed; project documentation at SpectrAI-Initiative.github.io/InnoClaw/.}
}

@misc{LabClaw2026,
  author = {Wu, Yingcheng Charles and Jian, Jinglin and Zhao, Zhe},
  title = {{LabClaw}: Operating layer for {LabOS} (Stanford-Princeton AI Co-Scientists)},
  year = {2026},
  howpublished = {GitHub repository},
  url = {https://github.com/wu-yc/LabClaw},
  note = {MIT licensed.}
}


\end{document}